\documentclass[11pt]{article}
\usepackage{iftex}
\ifPDFTeX
  \usepackage[utf8]{inputenc}
  \usepackage[T1]{fontenc}
  \usepackage{lmodern}
\else
  \usepackage{fontspec}
\fi
\usepackage[a4paper,margin=2.5cm]{geometry}
\usepackage{booktabs}
\usepackage{array}
\usepackage{url}
\usepackage[hidelinks]{hyperref}
\usepackage{microtype}
\ifPDFTeX \input{glyphtounicode}\pdfgentounicode=1 \fi

\title{Navital Gym: Rule-Graded Agent Environments\\for Indian Workflows}
\author{Daksh Bathla\\Navital Labs, Delhi, India\\\href{https://orcid.org/0009-0000-9085-6491}{ORCID: 0009-0000-9085-6491}}
\date{November 2026}

\begin{document}
\maketitle

\begin{abstract}
AI agents are being deployed on Indian workflows such as payments, tax, forms, travel and customer support in Hindi and Hinglish. Public agent evaluations rarely cover these workflows, and many score with another language model. We present Navital Gym, a set of six open environments with 321 public tasks in five task families: trip planning, UPI payments and disputes, GST invoicing and returns, form filling (including a multi-document split) and customer support in Hindi, Hinglish and three experimental Indian languages. Every task is graded by named, deterministic rules that recompute the outcome from a synthetic sandbox world. No language model judges any score. We report one run per task for three models on all six environments, Claude Opus 5.5, Claude Sonnet 5.5 and Claude Haiku 4.5, and for Gemini 3.1 Flash-Lite on two (UPI and the Support core). Opus solves 315 of 321 tasks, Sonnet 302 and Haiku 205. Trip planning and Hindi/Hinglish support separate the models most. We also describe a private held-out set that is not published, and a 7B adapter trained in the gym. Code is Apache-2.0 and task data is CC BY 4.0. The rules are a documented simplification of Indian practice, not the live systems.
\end{abstract}

\section{Introduction}

Agents that act for a user are being tried on tasks where the rules are written down and the consequences are real: a UPI payment that was debited but not credited, a GST invoice that must be computed exactly, a KYC form, a trip plan under rail quota and waitlist rules, a refund request written in Hinglish. These tasks have checkable answers. They also differ from the usual English-language tool-use tests in the details: Indian number formats, PAN and IFSC patterns, place-of-supply rules, code-mixed text.

We built Navital Gym to measure whether an agent gets such tasks right, with results that anyone can reproduce. Three design choices shape it.
\begin{enumerate}
\item \textbf{Written rules, no judge model.} A task is solved only if every applicable named rule passes. Each grader recomputes the result from the sandbox world and does not read what the agent claims about its own cost, score or success. Garbage input scores 0 and the grader does not crash.
\item \textbf{Honest refusals.} Many tasks have no valid answer. The agent must say so and give the right reason. A refusal with a wrong reason is marked wrong.
\item \textbf{Hidden fields stay hidden.} The agent sees the prompt, the tools and the policy text. Expected answers, feasibility labels and designer notes never reach it, and a test checks this for every task in every environment.
\end{enumerate}

\paragraph{Contributions.}
\begin{itemize}
\item Six open environments covering five task families, with 321 public tasks and simplified written rules with cited sources (Section~\ref{sec:envs}).
\item Graders built from independent oracles, with leak tests and scripted floor agents (Section~\ref{sec:valid}).
\item A public record of the grader and task fixes between v1.0.1 and v1.1, which changed scores (Section~\ref{sec:audit}).
\item One-run-per-task baselines for three models on all six environments and one on two, with pass@1, strict pass@1 and partial credit (Section~\ref{sec:results}).
\item A description of the private held-out set and of a small model trained in the gym (Sections~\ref{sec:heldout} and~\ref{sec:bay}), at the level of detail that is published today.
\end{itemize}

\paragraph{Scope.} The environments use synthetic data and a simplified version of Indian practice. They are tests for AI agents, not statements about current law, railway, bank or tax rules. Navital Gym is not affiliated with or endorsed by NPCI, RBI, GSTN, CBIC, Indian Railways, IRCTC or any bank.

\section{Related work}

\paragraph{Agent benchmarks with state-based grading.} $\tau$-bench~\cite{yao2024taubench} grades tool-using agents by comparing the final database state with an annotated goal state, and reports pass\textasciicircum$k$ for reliability across repeated trials. $\tau^2$-bench~\cite{barres2025tau2} extends this to a dual-control setting in which both the agent and a simulated user act on a shared environment. TravelPlanner~\cite{xie2024travelplanner} tests multi-constraint travel planning against a large sandbox of records. Navital Gym follows the same principle: the grader recomputes the outcome from the sandbox world and does not trust what the agent claims. Its rules are written and named for every task, and no language model judge enters the score.

\paragraph{Indian and Indic agent evaluation.} Labib and Yadati~\cite{labib2026secureagent} build three proof-of-concept Inspect environments for the Indian institutional context: government welfare eligibility, UPI dispute triage and IRCTC rail booking. Their abstract reports a deterministic oracle-based evaluator with per-step success over more than 300 instances, validation baselines on five LLMs, and a probe for how models respond to new rules when stale ones were seen before. It is a poster at the NeurIPS 2026 GlobalSouthAI affinity event, not a main-track paper. Two of their domains overlap with ours (UPI and rail booking), and the grading approach is similar.

In banking, TauIndianBankBench~\cite{adhikary2026tauindianbankbench} is an action-oriented benchmark of 1,000 Indian retail-banking tasks in which an assistant helps a simulated customer. Besides the final records it checks that the assistant took every required step and made no unrequested change. The authors report that for every model, 6.5 to 12.0\% of sessions are judged successful when only the final records are checked, although the assistant skipped a required step or made an unrequested change. The data is released on Hugging Face as \texttt{NPCI/tau-indian-banking}~\cite{npci2026taudata} (MIT licence; for the \texttt{indian\_banking} domain of the tau2-bench harness; 1,000 tasks across 50 scenario families, a synthetic bank database of 944 customers and a 24-document knowledge base; scored on the final database state and the required tool actions). NPCI also lists \texttt{NPCI/nemo-gym-indian-banking}~\cite{npci2026nemogym} (Apache-2.0; 300 synthetic multi-turn banking tasks, 250 train and 50 validation, in NVIDIA NeMo Gym format). IndicBankBench~\cite{paul2026indicbankbench} evaluates the safety and reliability of language models in Indian retail banking; its dataset \texttt{NPCI/IndicBankBench}~\cite{npci2026indicbankdata} holds 799 synthetic multi-turn cases under CC BY 4.0. Banking is not a Navital Gym task family. Its UPI environment is the closest overlap with these works, and the point made by TauIndianBankBench, that a final-state check can miss a skipped required step, is a limit our final-state-plus-forbidden-action rules share in part (Section~\ref{sec:limits}).

VoiceAgentBench~\cite{jain2025voiceagentbench} evaluates speech-based agents on single- and multi-tool calls, multi-turn dialogue and safety, in English and six Indic languages. Mukherjee, Bali and Sitaram~\cite{mukherjee2026actions} study how well tool-using agents retain policy across languages. Kamath et al.~\cite{kamath2025hindi} present a suite of datasets for benchmarking Hindi language models, and Inspect India Evals~\cite{singh2026inspectindia} is an open framework for evaluating language models in the Indian linguistic and cultural context.

\paragraph{Environments and verifiable rewards for tool use.} Reinforcement learning from verifiable rewards has been applied to software-engineering agents~\cite{da2025agentrlvr} and to enterprise tool use on Atlassian workflows~\cite{vissa2026atlassianrlvr}. Sarvam AI reports reinforcement learning, including tool use, for its open 30B and 105B models~\cite{sarvam2026models}.

\paragraph{Positioning.} Navital Gym combines open environments across five Indian task families, written named rules, and a private held-out set. Prior work covers parts of this space: banking~\cite{adhikary2026tauindianbankbench,paul2026indicbankbench}, UPI disputes and rail booking~\cite{labib2026secureagent}, speech agents in Indic languages~\cite{jain2025voiceagentbench} and cross-lingual tool-use policy~\cite{mukherjee2026actions}. We build on these and make no claim of priority. Our search was done on 3 November 2026 and is not exhaustive; the date is stated so that a reader can judge how current this section is.

\section{Environments and graders}\label{sec:envs}

\subsection{Shared core}
Every environment follows the same interface: \texttt{reset}, \texttt{step}, tools, a policy text, and \texttt{verify\_submission}. The core and all environments run on the Python 3.10+ standard library. Adapters for Inspect AI, verifiers and OpenEnv are optional extras. They have offline end-to-end tests with a scripted model but have not been run in a container or a hosted space. Any model with an OpenAI-compatible chat API and tool calling can be run, including local servers. Pass@1 uses first-run semantics: the first attempt on each task counts.

\subsection{The six environments}

\begin{table}[h]
\centering\small
\begin{tabular}{@{}l p{7.4cm} r@{}}
\toprule
env & what the agent does & tasks\\
\midrule
Trip & Indian trip planning under real constraints: trains with quotas and waitlists, flights, road, hotels, seasons, permits & 55 (17 infeasible)\\
UPI & payment attempts and disputes under written, simplified rules & 50\\
GST & invoicing and return computations from documents & 50\\
Form & filling and checking forms (KYC-style fields, addresses, numbers) & 50\\
Form multi-document & cross-checking identity documents and records; reporting what disagrees & 40\\
Support & e-commerce returns, refunds and privacy requests in Hindi and Hinglish (core), with Tamil, Bengali and Marathi as a labelled experimental split & 50 core + 26 experimental\\
\bottomrule
\end{tabular}
\caption{The six environments. Total: $55+50+50+50+40+76=321$ tasks.}
\end{table}

Form and Form multi-document are two environments of one task family (form filling), so the six environments make five task families.

\paragraph{Trip} has 25 cities, 41 links, trains with IRCTC-style classes and quotas (General, Tatkal and Current), seat status of AVL, RAC and WL, flights, taxis and buses with seasonal road closures, hotels and 48 sights with closing days, hours, seasons and permits. The agent is told all 17 rules, for example R-QUOTA, R-AVAIL, R-OPEN, R-BUD and R-PERMIT. A plan is solved only if every applicable rule passes. Tatkal booking needs the traveller's Aadhaar OTP, so the agent can only mark \texttt{traveller\_otp} and cannot complete the booking. The data is synthetic and not affiliated with IRCTC.

\paragraph{UPI} has 50 tasks in seven kinds: pay after checking the payee's name (10), payment limits (8), money debited for a failed payment (10), a payment sent to the wrong VPA (5), collect requests, some of them scams (7), AutoPay mandates (5) and UPI Lite (5). Rules were last checked on 25 October 2026. Seven of the tasks are collect requests: six are scam requests from a fake merchant VPA that the agent should decline and report, and one is a genuine merchant request that it should approve. They are kept because they test scam handling. The documented rules record only that person-to-person collect ended on 1 October 2025; the later NPCI change to merchant collect described in Section~\ref{sec:bay} is not modelled in these public tasks.

\paragraph{GST} gives a short accounting question; the agent reads documents and the policy through tools, does exact decimal arithmetic and returns a structured answer. The rules are a deliberately simplified version of Indian GST with no QRMP, state-specific rules, reverse charge or cess.

\paragraph{Form} includes KYC fields from messy document text (15 of 50 tasks), address parsing (12), PIN-to-state checks (8), Indian number idioms such as ``2.5 lakh'' (10) and fiscal-year dates (5). \textbf{Form multi-document} has five families of 8 tasks each (multi-document disagreement, PAN-name structure, record conflicts, OCR repair, batch) and is reported separately because it is an experimental harder split.

\paragraph{Support} is an invented online shop with an invented policy. The core has 25 Hindi and 25 Hinglish tasks; the experimental split has Tamil 8, Bengali 9 and Marathi 9. The reply language is a hard rule. Hindi lines were machine-checked with a Sarvam model and Hinglish lines were drafted with AI assistance; neither was reviewed by a native speaker. The Tamil, Bengali and Marathi text was machine-assisted and kept when a back-translation matched the intended meaning, which is a consistency check and not a review. The author's read of the Hinglish lines is still pending, so we make no native-review claim for them.

\subsection{Rules and grader validity}\label{sec:valid}
Each environment has named rules. Examples from Support: R-ACTIONS (every required action is present with the exact amount and mode), R-FORBIDDEN (no extra action, such as a double refund), R-DECISION and R-REASON (a structured decision and reason code graded against an independent oracle, so declining without reading the order cannot pass), R-AMOUNT (the reply states the refund and no other rupee figure), R-PRIVACY (no other customer's address or phone) and R-LANG (the reply is in the customer's language).

Three checks support the validity of the graders.
\begin{enumerate}
\item \textbf{Oracle agents.} Each environment has a scripted reference agent that should score 1.00. For Form multi-document, the oracle agrees with the generator on all 40 tasks and a reviewer hand-solved all 40 independently and agreed.
\item \textbf{Scripted floor agents.} Agents that read nothing (empty, always-refuse, majority-answer) should not solve more than 25\% of tasks. On the 50 UPI tasks they solve 0, 0 and 2. On Support, the majority-vote agent solves 13\% of the 76 tasks and an agent that gives the right actions with an English reply solves 0\%. \textbf{The 25\% floor does not hold on the public GST set:} a built-in \texttt{careless} baseline that reads everything but makes two slips scores 80\% there. The repository's results note says the gate holds only on the private held-out sets. We report no held-out scores here.
\item \textbf{Leak tests.} A test runs a leak check over every task in every environment, so hidden fields do not reach the agent.
\end{enumerate}

\subsection{What changed from v1.0.1 to v1.1}\label{sec:audit}
After v1.0.0 and v1.0.1 we found and fixed errors in our own graders and tasks, and re-graded all rows. v1.1 scores are not comparable with v1.0.1 scores, so both are shown. The main changes were:
\begin{itemize}
\item Trip: a Ladakh fee rule for nights in Nubra or Pangong, the Tatkal Aadhaar OTP hand-off, a general quota window of 1 to 60 days, return city and origin stated in prompts, exact arithmetic in the budget rule.
\item UPI: mandate prompts say registration needs the owner's own PIN; amounts are compared exactly.
\item Support: refund amounts checked exactly, wider privacy checks, and a reply-quality floor on the language rule.
\item Core: a rule that raises is reported by name; a tool error is returned to the agent instead of crashing the run.
\end{itemize}
For the three Claude models the effect on all six environments was:

\begin{table}[h]
\centering\small
\begin{tabular}{@{}lrr@{}}
\toprule
model & v1.0.1 & v1.1\\
\midrule
Claude Opus 5.5 & 264/321 & 315/321\\
Claude Sonnet 5.5 & 256/321 & 302/321\\
Claude Haiku 4.5 & 177/321 & 205/321\\
\bottomrule
\end{tabular}
\caption{Tasks solved, all six environments, v1.0.1 and v1.1 (changelog of the public repository).}
\end{table}

Grader and task bugs move scores by tens of tasks. The changelog also lists known issues we have not fixed, for example that some ways of stating a refund in the Support R-AMOUNT rule are not checked.

\section{Held-out tasks and generators}\label{sec:heldout}
Public tasks can leak into training data. For that reason Navital Labs keeps a private held-out set of tasks that has never been published, and has generators that make new tasks for five of the six task sets. Navital Labs states on its website that the private held-out set holds 181 tasks, behind the 321 public ones. We give the count only: this paper reports no held-out scores and does not describe how the set was built.

The private held-out set is run only on our own compute and is not sent to free hosted APIs; this is our practice, not a technical guarantee. This set is separate from the UPI test set used for Navital Bay in Section~\ref{sec:bay}, which we call the \emph{Bay UPI test set} and which we did run through the Anthropic API. Private evaluations are a paid service of Navital Labs (Section~\ref{sec:limits}). The public tasks can be run by anyone.

\section{Results on the public set}\label{sec:results}

\paragraph{Setup.} Every number below is pass@1 with $k=1$: one run per task, at most 40 turns, one format-repair retry. The saved settings of all 20 run files (the 19 cells of Table~\ref{tab:main} and the Gemini Support run) record temperature 0.0, 40 turns and repair retry on; Anthropic rows also record a 4,096-token output limit. These settings are the requested ones. The harness drops the temperature parameter and carries on if the API rejects it, and the run files do not record whether that happened. A later probe found that Claude Sonnet 5.5 and Opus 5.5 reject the parameter (HTTP 400, ``temperature is deprecated for this model''), so their rows ran at the API's default temperature, not 0; Claude Haiku 4.5 accepted temperature 0. The scores are as run. We have not probed Gemini 3.1 Flash-Lite on this point. All rows were re-graded by the same grading code at grader tree \texttt{1543b5103be5}. There are 321 tasks in all.

\begin{table}[h]
\centering\small
\setlength{\tabcolsep}{4pt}
\begin{tabular}{@{}lrrrrrrr@{}}
\toprule
model & Trip & UPI & GST & Form & Form md & Support & six envs\\
\midrule
Claude Opus 5.5 & 51/55 & 50/50 & 50/50 & 50/50 & 40/40 & 74/76 & 315/321\\
Claude Sonnet 5.5 & 46/55 & 47/50 & 50/50 & 50/50 & 40/40 & 69/76 & 302/321\\
Claude Haiku 4.5 & 19/55 & 26/50 & 42/50 & 47/50 & 39/40 & 32/76 & 205/321\\
Gemini 3.1 Flash-Lite & -- & 39/50 & -- & -- & -- & -- & not run on all six\\
\bottomrule
\end{tabular}
\caption{Tasks solved out of tasks, v1.1, $k=1$. ``Form md'' is Form multi-document. Six-environment totals as a share of 321: Opus 98.1\% (95\% Wilson interval 96--99), Sonnet 94.1\% (91--96), Haiku 63.9\% (58--69). The intervals are computed from the counts.}
\label{tab:main}
\end{table}

Gemini 3.1 Flash-Lite was run on the UPI full set and on the 50 Support core tasks only (21/50, not shown in the table because the column is the full 76-task split), because of free-tier quota. Its partial Support splits are excluded from the public leaderboard with the reason stated.

\paragraph{Known issue beside Table~\ref{tab:main} (UPI answer layout).} The v1.1 UPI instructions did not say that the answer must be a flat object. Some models put the detail fields under a label key, and the grader, which reads the top level only, marked them missing even when the values were right. The official scores stay as published. When one level of nesting is accepted and nothing else changes, Claude Haiku 4.5 on UPI goes from 26 to 32 of 50 (total 211 of 321); no other model in this paper changes, and the order of the models does not change. The v1.2 instructions will state a flat schema, and every model will be re-run (public changelog, ``Known issues (fixed in v1.2)'').

\paragraph{What the table shows.} With one run per task, small differences are not meaningful: at $k=1$ an 8-task gap does not separate two models. We therefore make three claims only.
\begin{enumerate}
\item Form, Form multi-document and GST are close to saturated for the two largest Claude models (50 + 50 + 40 = 140 tasks, and both Opus and Sonnet solve all 140), so these sets separate weaker models but not the top.
\item Trip planning and Support separate the models most. On Trip, solved counts range from 19 to 51 of 55; on Support from 32 to 74 of 76.
\item The experimental Support split (Tamil, Bengali, Marathi) is small: 26 tasks. Opus solves 26, Sonnet 24 and Haiku 8; the core split gives 48, 45 and 24 of 50. We do not rank models on 26 tasks, and the split is machine-assisted and not native-reviewed.
\end{enumerate}

\paragraph{Strict pass@1 and partial credit.} Strict pass@1 counts a task only if the answer was valid without the one format-repair retry. For Trip, Opus solves 92.7\% of tasks and 67.3\% strictly, Sonnet 83.6\% and 54.5\%, and Haiku 34.5\% and 32.7\%. For Haiku the partial-credit score (mean fraction of scored rules passed) shows progress where pass@1 is low: on Trip it is 0.66. The repair retry changes Trip scores a lot and should be reported with any Trip number.

\paragraph{Length cut-offs.} The public leaderboard export counts, for each row, the unsolved tasks whose first attempt hit the model's output limit before a final answer; a task that the format-repair retry still solved is not counted. Every row in Table~\ref{tab:main} has 0.

\paragraph{Rows not re-run.} The changelog lists every v1.0.1 row that has no v1.1 re-run, for example GPT-4o mini and gpt-oss easy-tier runs. They are not in Table~\ref{tab:main} and are not comparable with it.

\section{A small model trained in the gym: Navital Bay}\label{sec:bay}

We trained Navital Bay, a rank-16 LoRA adapter on Qwen2.5-7B-Instruct for UPI tasks, using worked examples from the gym. The adapter is on Hugging Face under Apache-2.0; merged weights are not published~\cite{navitalbay2026}.

The following is what the model card reports as of 9 November 2026. We call its test tasks the \emph{Bay UPI test set}. They are new UPI tasks for the three flows that are current today (failed-but-debited, pay-and-verify, UPI Lite), 75 per kind, generated after the training data was locked; none appeared in training. They are graded by rules, not by a model. The set is separate from the private held-out set of Section~\ref{sec:heldout}: it was run on the Anthropic API for the two Claude rows, which we do not do with the private set. Each model ran once per task. Temperature 0 was used except for Claude Sonnet 5.5, which does not accept the setting and ran at the API's default. Output limits were 4,096 tokens for the Anthropic models and 1,500 for our local 7B models.

\begin{table}[h]
\centering\small
\begin{tabular}{@{}lrrrr@{}}
\toprule
model & failed but debited & pay and verify & UPI Lite & forbidden attempts\\
 & (of 75) & (of 75) & (of 75) & (of 118)\\
\midrule
Navital Bay (7B) & 73 & 75 & 61 & 3\\
Claude Sonnet 5.5 & 75 & 75 & 75 & 0\\
Claude Haiku 4.5 & 0 & 65 & 75 & 36\\
Previous training run (7B) & 63 & 45 & 48 & 7\\
\bottomrule
\end{tabular}
\caption{Bay UPI test set; numbers as of 9 November 2026, as on the model card. The 118 are the tasks, out of the 225 on these three flows, in which the rules forbid at least one action; in the other 107 no forbidden action was possible.}
\end{table}

Claude Haiku 4.5 scored 0 on failed-but-debited. In the run file 74 of its 75 answers put the detail fields under a label key, which the grader does not read; this is the same layout problem as in Section~\ref{sec:results}. Many of these answers also had a wrong detail value or broke other rules, and we did not re-grade this set with nesting accepted, so we do not say how many it would solve. A forbidden attempt counts a try at an action the rules forbid, even when the simulated bank would have refused it. Claude Sonnet 5.5 scores higher than Navital Bay on failed-but-debited and UPI Lite and ties on pay-and-verify. Differences of a few tasks may not be meaningful at one run each. The model was trained on these kinds of task and is not a test of how well it generalises. It is not built for GST, and UPI mandates were not part of its training. Collect-request tasks are not counted in these results, because NPCI phased out most merchant collect payments from 28 February 2026.

\section{Limitations, conflict of interest and ethics}\label{sec:limits}

\paragraph{Limitations.}
\begin{itemize}
\item The rules are a documented simplification and can be out of date. Rules were last checked on 25 October 2026 (UPI, Support) and 31 October 2026 (Trip, GST, Form). The Income-tax Act 2025 and the Tatkal OTP step were missing from the v1.0.1 rules and were added in v1.1.
\item Small $n$ and $k=1$ in most cells. One task is 1.3 to 2.5 points of a cell of 40 to 76 tasks. The six-environment totals have 95\% Wilson intervals that are 3 to 11 points wide.
\item The Trip pass@1 depends on one format-repair retry.
\item The Tamil, Bengali and Marathi split and the Hinglish lines are machine-assisted and not native-reviewed. The language check is a script-and-marker-word heuristic, and a code-mixed reply below 70\% in the customer's script fails by design.
\item Known grader gaps in the Support R-AMOUNT rule. A final-state-plus-forbidden-action grader can miss a required step that leaves no trace, a risk that the TauIndianBankBench abstract describes for final-record-only checks~\cite{adhikary2026tauindianbankbench}.
\item The Inspect, verifiers and OpenEnv adapters have not been run in containers or hosted spaces.
\item Models not re-run for v1.1 are listed with their v1.0.1 results only.
\item The models in Table~\ref{tab:main} are not the strongest agents available. The results show that the environments separate models. They are not a leaderboard of the field.
\end{itemize}

\paragraph{Conflict of interest.} The author runs Navital Labs, which sells private evaluations, and trains models in its environments and tests them on held-out tasks. The public benchmark and its results are open and can be re-run. Navital Labs is in Anthropic's Claude for Startups programme and the Sarvam Startup Program, and its API use includes credits from Anthropic and from Sarvam.

\paragraph{Ethics and data.} Code is Apache-2.0 and task data is CC BY 4.0. All people, accounts, PANs and GSTINs are synthetic. GSTINs have a valid check character on fake PAN-shaped bodies and do not belong to real taxpayers; no Aadhaar numbers are used. The environments are not financial, tax or travel advice.

\section*{Acknowledgements}
The author used Claude (Anthropic) as a writing and coding assistant. The author designed the study and the environments, directed the work, and reviewed all text and references, and takes full responsibility for the content. No AI tool is an author.

\appendix
\section{Reproduction}
Install with \texttt{pip install navital-gym}, then for example \texttt{navital-gym run --env trip --agent openai:<model>@<base\_url>}. To reproduce every row in Table~\ref{tab:main}, use tag \texttt{v1.1.0} or the public \texttt{main} branch at commit \texttt{30922a7} of the repository~\cite{navitalgym2026}, which carry the same rows for these models. To verify the grader, \texttt{git rev-parse HEAD:navital\_gym} starts with \texttt{1543b5103be5} in both. Per-run pass@1, strict pass@1 and partial credit for every model and environment are in \texttt{docs/leaderboard.json} and \texttt{docs/RESULTS.md} of that repository.

\bibliographystyle{plain}
\bibliography{refs}

\end{document}